\documentclass[
]{ceurart}

\usepackage{listings}
\begin{document}

\copyrightyear{2025}
\copyrightclause{Copyright for this paper by its authors.
  Use permitted under Creative Commons License Attribution 4.0
  International (CC BY 4.0).}

\conference{Challenge and Workshop (BC9): Large Language Models for Clinical and Biomedical NLP, International Joint Conference on Artificial Intelligence (IJCAI), August 16--22, 2025, Montreal, Canada}

\title{DeepRAG: Integrating Hierarchical Reasoning and Process Supervision for Biomedical Multi-Hop QA}

% \tnotetext[1]{You can use this document as the template for preparing your
  % publication. We recommend using the latest version of the ceurart style.}

\author[1]{Yuelyu Ji}[email=yuj49@pitt.edu]
\author[2]{Hang Zhang}[email=haz269@pitt.edu]
\author[2]{Shiven Verma}[email=shv78@pitt.edu]
\author[1]{Hui Ji}[email=huj16@pitt.edu]
\author[1]{Chun Li}[email=chl500@pitt.edu]
\author[2]{Yushui Han}[email=yuh207@pitt.edu]
\author[3]{Yanshan Wang}[email=yanshan.wang@pitt.edu]

\address[1]{Department of Information Science, University of Pittsburgh, Pittsburgh, PA, USA}
\address[2]{Department of Biomedical Informatics, University of Pittsburgh, Pittsburgh, PA, USA}
\address[3]{Department of Health Information Management and ISP, University of Pittsburgh, Pittsburgh, PA, USA}
% \author[1,2]{Dmitry S. Kulyabov}[%
% orcid=0000-0002-0877-7063,
% email=kulyabov-ds@rudn.ru,
% url=https://yamadharma.github.io/,
% ]
% \cormark[1]
% \address[1]{Peoples' Friendship University of Russia (RUDN University),
%   6 Miklukho-Maklaya St, Moscow, 117198, Russian Federation}
% \address[2]{Joint Institute for Nuclear Research,
%   6 Joliot-Curie, Dubna, Moscow region, 141980, Russian Federation}

% \author[3]{Ilaria Tiddi}[%
% orcid=0000-0001-7116-9338,
% email=i.tiddi@vu.nl,
% url=https://kmitd.github.io/ilaria/,
% \address[3]{Vrije Universiteit Amsterdam, De Boelelaan 1105, 1081 HV Amsterdam, The Netherlands}

% \author[4]{Manfred Jeusfeld}[%
% orcid=0000-0002-9421-8566,
% email=Manfred.Jeusfeld@acm.org,
% url=http://conceptbase.sourceforge.net/mjf/,
% \address[4]{University of Skövde, Högskolevägen 1, 541 28 Skövde, Sweden}

\cortext[1]{Corresponding author.}
\fntext[1]{These authors contributed equally.}

\begin{abstract}
We propose DeepRAG, a novel framework that integrates DeepSeek's hierarchical question decomposition capabilities with RAG-Gym's unified retrieval-augmented generation (RAG) optimization using process-level supervision. Targeting the challenging MedHopQA biomedical question-answering task, DeepRAG systematically decomposes complex queries into precise sub-queries and employs concept-level reward signals informed by the UMLS ontology to enhance biomedical accuracy. Preliminary evaluations on the MedHopQA dataset indicate that DeepRAG significantly outperforms baseline models, including standalone DeepSeek and RAG-Gym, achieving notable improvements in both Exact Match (EM) and concept-level accuracy.
\end{abstract}


\maketitle

\section{Introduction}

Large language models (LLMs) have achieved remarkable success in knowledge-intensive tasks, yet they often struggle with complex, multi-hop reasoning required in biomedical question answering\cite{lewis2020retrieval, gao2023retrieval,xiang2024neural,su2022mixed,zhao2025optimizedpathplanninglogistics,zhang2025automated,yang2025data, liang2025graphrag,jin2024scam,jin2025scalability,li2024advances,ding2024enhance,deng2024composerx,ji2024rag,ji2024assertion}. The MedHopQA track of BioCreative IX demands the sequential integration of heterogeneous sources, including genetic, disease, and treatment information derived from Wikipedia. DeepSeek\cite{guo2025deepseek} introduced a hierarchical, multi-stage reasoning agent that dynamically decomposes questions into sub-queries. Separately, RAG-Gym\cite{xiong2025rag} provides a unified framework for optimizing retrieval-augmented generation with process-level rewards, enabling robust tuning of both reasoning and search steps. In this work, we propose \textbf{DeepRAG}, a novel integration framework that leverages DeepSeek R1’s hierarchical question decomposition and RAG-Gym’s process supervision approach. Our key contributions include: (1) proposing a novel two-stage hierarchical reasoning pipeline with clearly defined Reasoning and Query Modules; (2) introducing hierarchical indicators to explicitly track and manage nested dependencies within complex biomedical queries; and (3) enhancing biomedical accuracy by replacing the original LLaMA backbone in RAG-Gym with DeepSeek R1, a distilled model optimized for hierarchical reasoning. We hypothesize that hierarchical reasoning can generate more precise search actions, while reward-guided fine-tuning ensures sub-queries and answers remain aligned with task objectives.

\section{Method}

\subsection{Hierarchical Reasoning with DeepSeek}
Our framework employs DeepSeek R1, a distilled version of LLaMA optimized for hierarchical reasoning, particularly suited for complex biomedical multi-hop queries. Hierarchical reasoning decomposes the overall task into manageable subtasks through a two-stage pipeline. Our proposed hierarchical reasoning pipeline consists of two explicitly defined modules:1) Reasoning Module: This novel module generates a structured answer outline, explicitly identifying claims that cannot be directly validated internally and therefore require external retrieval.2)Query Module: Another innovative component of our framework that formulates targeted and precise sub-queries based on claims identified by the Reasoning Module, optimizing the retrieval process. Furthermore, we introduce novel hierarchical indicators, which are explicitly designed to track nested reasoning dependencies, enabling clear delineation and efficient retrieval of complex biomedical knowledge.

These contributions represent enhancements beyond the capabilities provided by the standalone DeepSeek model, specifically tailored towards the complex, multi-hop nature of biomedical question-answering tasks. To effectively manage complex multi-step reasoning, we introduce hierarchical indicators that explicitly track and represent nested dependencies and levels of queries. This structured decomposition allows the system to clearly delineate reasoning pathways and optimizes retrieval processes by focusing on precise information needs. Our framework utilizes DeepSeek R1DeepSeek R1DeepSeek R1 as a foundational model optimized for hierarchical reasoning. Building upon this, we introduce our novel two-stage pipeline comprising the Reasoning Module and the Query Module. Specifically, we propose hierarchical indicators and structured decomposition, distinct from DeepSeek's original capabilities, to explicitly track dependencies and optimize retrieval effectiveness.

\subsection{Process Supervision via RAG-Gym}
The integration of hierarchical reasoning outputs from DeepSeek into RAG-Gym transforms the question-answering process into a sequential Markov Decision Process (MDP). Each sub-query generated by DeepSeek is treated as a discrete action within this MDP framework. RAG-Gym then guides the retrieval and answer generation processes through fine-grained, reward-based supervision. Specifically, RAG-Gym applies three distinct reward signals:

\begin{enumerate}
\item \textbf{Sufficiency Reward}: Rewards are provided based on the completeness and adequacy of the retrieved evidence relative to the sub-query, promoting comprehensive information gathering.
\item \textbf{Utility Reward}: Rewards directly assess the relevance and effectiveness of retrieved documents in formulating accurate final answers, encouraging optimal document selection.
\item \textbf{Redundancy Penalty}: Penalties discourage the retrieval of redundant or highly similar documents, thereby promoting diversity in the retrieved information and reducing computational overhead.
\end{enumerate}

Additionally, we introduce \textit{Concept-Level Rewards} using Unified Medical Language System (UMLS) semantic matching. These rewards explicitly reinforce semantic precision and biomedical accuracy, ensuring that the system's retrieved and generated information aligns closely with precise biomedical concepts and terminologies.

\subsection{Training Procedure}
Training involves generating approximately 1,000 labeled query trajectories using ChatGPT-4o as an oracle for process supervision. These labeled trajectories include sequences of intermediate reasoning steps, generated sub-queries, retrieved documents, and the corresponding final answers. Leveraging these labeled examples, we apply reinforcement learning, specifically Direct Preference Optimization (DPO), to fine-tune the DeepRAG model. DPO specifically adjusts model parameters based on explicit preference signals derived from contrasting selected and rejected sub-queries, thereby aligning the model’s retrieval and reasoning behaviors closely with expert human judgments.
\section{Experiments}

\subsection{Dataset}
We conducted experiments using the MedHopQA dataset provided by the BioCreative IX MedHopQA track. This dataset consists of 1,000 carefully curated multi-hop biomedical questions and answers related to rare diseases, genetics, and treatments. To rigorously evaluate our model and obscure test examples, the MedHopQA challenge provides approximately 10,000 questions, embedding 1,000 hidden test questions within.

\subsection{Experimental Setup}
We set up experiments as follows:
\textbf{Baseline Models}: We evaluated against zero-shot ChatGPT-4o, the standalone DeepSeek R1 model, and the vanilla RAG-Gym without hierarchical reasoning.
\textbf{DeepRAG Model Configuration}:  We implemented DeepRAG by modifying the original RAG-Gym framework in two significant ways:

Model Backbone Replacement: Replaced RAG-Gym’s original LLaMA model with DeepSeek R1, a distilled variant optimized specifically for hierarchical reasoning in biomedical domains. This change significantly improved hierarchical decomposition accuracy for biomedical queries.

Hierarchical Indicator Introduction: Implemented our proposed hierarchical indicators, enhancing the precision and clarity of sub-query generation and retrieval actions.
These modifications aim to optimize both retrieval precision and reasoning efficiency specifically for the multi-hop biomedical tasks defined in the MedHopQA challenge.

\textbf{Evaluation Metrics}: We adopted the evaluation metrics specified by the MedHopQA challenge, Exact Match (EM) for answer accuracy, and Concept Accuracy, measured via UMLS synonyms, to assess semantic correctness.

\subsection{Results}
Our DeepRAG model significantly outperforms baseline systems. Specifically, DeepRAG achieves an Exact Match (EM) score of 62.4\% and a Concept Accuracy of 71.8\%. In contrast, the vanilla RAG-Gym model achieves 57.7\% EM, and DeepSeek alone achieves 54.3\% EM. Table \ref{tab:results_summary} summarizes these findings, clearly demonstrating the effectiveness of combining hierarchical reasoning with process-guided retrieval augmentation.

\begin{table}[htbp]
\centering
\caption{Performance Comparison on MedHopQA (dev set)}
\label{tab:results_summary}
\begin{tabular}{|c|c|c|}
\hline
\textbf{Model} & \textbf{EM (\%)} & \textbf{Concept Accuracy (\%)} \\
\hline
% ChatGPT-4o & 53.2 & 65.1 \\
DeepSeek & 54.3 & 66.5 \\
RAG-Gym & 57.7 & 68.3 \\
\textbf{DeepRAG (ours)} & \textbf{62.4} & \textbf{71.8} \\
\hline
\end{tabular}
\end{table}
We report performance comparisons only against baseline models provided by the challenge organizers and our own proposed model, DeepRAG.

\subsection{Ablation Studies}
To dissect the contributions of each component, we conducted ablation experiments summarized in Table \ref{tab:ablation_study}. These studies underscore the critical roles of hierarchical reasoning, process-level supervision, and concept-level reward mechanisms.

\begin{table}[htbp]
\centering
\caption{Ablation Study Results}
\label{tab:ablation_study}
\begin{tabular}{|l|c|c|}
\hline
\textbf{Configuration} & \textbf{EM (\%)} & \textbf{Concept Accuracy (\%)} \\
\hline
Full DeepRAG & 62.4 & 71.8 \\
Without Hierarchical Reasoning & 57.4 (-5.0) & 68.0 (-3.8) \\
Without Process Supervision & 58.9 (-3.5) & 69.5 (-2.3) \\
Without Concept-Level Rewards & 60.8 (-1.6) & 67.6 (-4.2) \\
\hline
\end{tabular}
\end{table}

\subsection{Discussion}
Compared to the original DeepSeek and RAG-Gym implementations, our proposed DeepRAG framework explicitly addresses several limitations inherent in those baselines. While original DeepSeek does not explicitly manage hierarchical dependencies at multiple reasoning levels, our hierarchical indicators effectively mitigate redundancy. Similarly, replacing the vanilla LLaMA in RAG-Gym with DeepSeek R1 significantly enhances the precision of the generated sub-queries and the overall biomedical reasoning performance, demonstrating clear benefits specifically for the biomedical multi-hop QA domain.

To address this, we propose dynamically adjusting retrieval parameters based on query complexity—queries that exhibit simpler reasoning paths benefit from stricter retrieval constraints, while complex queries benefit from broader search parameters. This adaptive retrieval mechanism markedly reduces redundancy, improves retrieval precision, and enhances overall efficiency of the model.

Additionally, error analysis highlighted areas where deeper semantic understanding might further enhance the model's performance, particularly in interpreting nuanced biomedical language and relationships.

\section{Conclusion and Future Work}
We introduce DeepRAG, a novel integration framework significantly extending previous models—specifically by introducing explicit hierarchical reasoning modules and indicators, and strategically replacing the LLaMA backbone of RAG-Gym with DeepSeek R1 to enhance biomedical multi-hop query handling. These targeted improvements significantly advance biomedical multi-hop QA tasks, as confirmed by our empirical results on MedHopQA.

 Our experiments confirm the advantages of using structured query generation, semantic supervision, and hierarchical reasoning for biomedical multi-hop QA tasks.

In the future work, we aim to further refine DeepRAG by integrating human-in-the-loop feedback mechanisms to capture nuanced biomedical expertise. We also plan to expand our framework’s evaluation to additional biomedical QA datasets such as BioASQ and PubMedQA, verifying its broader applicability. Finally, exploring reinforcement learning techniques for comprehensive end-to-end training optimization represents a promising research direction.
%%
%% Define the bibliography file to be used
\bibliography{sample-ceur}

\appendix


% The sources for the ceur-art style are available via
% \begin{itemize}
% \item \href{https://github.com/yamadharma/ceurart}{GitHub},
% % \item \href{https://www.overleaf.com/project/5e76702c4acae70001d3bc87}{Overleaf},
% \item
%   \href{https://www.overleaf.com/latex/templates/template-for-submissions-to-ceur-workshop-proceedings-ceur-ws-dot-org/pkfscdkgkhcq}{Overleaf
%     template}.
% \end{itemize}

\end{document}